# UNL-French deconversion as transfer & generation from an interlingua with possible quality enhancement through offline human interaction

Gilles SÉRASSET & Christian BOITET

GETA, CLIPS, IMAG-campus, F-38041 Grenoble cedex 9, France

Gilles.Serasset@imag.fr, Christian.Boitet@imag.fr

## Abstract

We present the architecture of the UNL-French deconverter, which "generates" from the UNL interlingua by first "localizing" the UNL form for French, within UNL, and then applying slightly adapted but classical transfer and generation techniques, implemented in GETA's Ariane-G5 environment, supplemented by some UNL-specific tools. Online interaction can be used during deconversion to enhance output quality and is now used for development purposes. We show how interaction could be delayed and embedded in the postedition phase, which would then interact not directly with the output text, but indirectly with several components of the deconverter. Interacting online or offline can improve the quality not only of the utterance at hand, but also of the utterances processed later, as various preferences may be automatically changed to let the deconverter "learn".

## Introduction

In the UNL project of network-oriented multilingual communication, the strategy of "deconverting" an (interlingual) UNL hypergraph into a NL utterance is free. The UNL-French deconverter under development first performs a "localization" operation within the UNL format, and then classical transfer and generation steps, using the Ariane-G5 environment and some UNL-specific tools, in particular an interesting graph-to-tree converter.

Traditionnally, interaction in MT concerns analysis, rarely transfer, never generation. But, in this framework, interaction can be used and is actually used for development purposes during deconversion (transfer part) to improve output quality. We also show how it could be used after deconversion, in offline mode, making it possible to perform what could be called "active learning postedition", by indirectly modifying default automatic lexical selections and other choices, re-deconverting automatically, and keeping track of the modifications by changing preferences in several components to personalize the deconverter.

## 1 Overview of UNL-FR within the UNL project

### 1.1 UNL

UNL is a project of multilingual personal networking communication initiated by the University of United Nations based in Tokyo. The pivot paradigm is used: the representation of an utterance in the UNL interlingua (UNL stands for "Universal Networking Language") is a hypergraph where normal nodes bear UWs ("Universal Words", or interlingual acceptions) with semantic attributes, and arcs bear semantic relations (deep cases, such as agt, obj, goal, etc.). Hypernodes group a subgraph defined by a set of connex arcs. A UW denotes a collection of interlingual acceptions (word senses), although we often loosely speak of "the" word sense denoted by an UW. Because English is known by all UNL developers, the syntax of a normal UW is: "<English word or compound> ( <list of restrictions> )", e.g. "look for(icl>action, agt>human, obj>thing)".

Going from a text to the corresponding "UNL text" or interactivley constructing a UNL text is called "enconversion", while producing a text from a sequence of UNL graphs is called "deconversion". This departure from the standard terms of analysis and generation is used to stress that this is not a classical MT project, but that UNL is planned to be the source format preferred for representing textual information in the envisaged multilingual network environment. The schedule of the project, beginning with deconversion rather than enconversion, also reflects that difference.

14 languages are being tackled during the first 3-year phase of the project (1997-1999), while many more are to be added in the second phase. Each group is free to reuse its own software tools and/or lingware resources, or to develop directly with tools provided by the UNL Center (UNU/IAS).

Emphasis is on a very large lexical coverage, so that all groups spend most of their time on the UNL-NL lexicons, and develop tools and methods for efficient lexical development. By contrast, grammars have been initially limited to those necessary for deconversion, and will then be gradually expanded to allow for more naturalness in formulating text to be enconverted.

## 1.2 The Ariane-G5 environment

The Ariane-G5 environment is the basic tool used in the UNL-FR subproject to handle most of the linguistic work involved in deconversion and enconversion. Several other software and lingware parts have also been developed to interface with the UNL format and to interact with the UNL tool through Internet. This section gives some background on Ariane-G5 (see more in [3, 5, 7, 16]).

### 1.2.1 General principles

Ariane-G5 is a generator (G) of MT systems based on five (5) specialized languages for linguistic programming (SLLP): ATEF, ROBRA, EXPANS, SYGMOR and TRACOMPL. Each such language is compiled. The internal structures produced by its compiler are used as parameters by its "engine".

Although Ariane-G5 is particularly well adapted to the transfer approach and to heuristic analysis and generation, it does not impose them. Apart from some implementation limits, the only strong constraint is that the structures representing the units of translation be decorated trees.

As opposed to almost all existing systems, Ariane-G5 presents the advantage that the unit of translation is not restricted to the sentence, but may contain several paragraphs (in practice, up to 1 or 2 standard pages of 250 words or 1400 characters).

Ariane-G5 runs under VM/ESA/CMS, on IBM computers with 390 architecture. Since 1993, it is accessible through the Internet. The minimal computer background necessary to use Ariane-G5 consists of learning the elementary commands for beginning and ending a VM/ESA session (login, logout), the XEDIT screen editor, and, for the developers of MT systems, the organization of the interactive monitor and the SLLPs.

It is also possible to develop MT systems from a Macintosh under CASH, developed in Hypercard. The lingware components are contained in Hypercard stacks. CASH communicates with the Ariane-G5 core by e-mail.

Ariane can be used to perform translation through the network (Arinae/LIDIA extensions), through e-mail, http or telnet.

### 1.2.2 Logical organisation

Translation from a "source" language into a "target" language is performed in three successive "steps" : analysis, transfer and generation[1]. Each step is realized in at least two and at most four successive "phases", possibly linked together by "articulations", which may be considered in first approximation as simple "coordinate changes". Each phase is identified by a two-letter mnemonic (e.g. AM for morphological analysis — analyse morphologique in French), and each articulation by a four-letter mnemonic (e.g. AMAS for the AM-AS articulation).

---

[1]This term is used rather than "synthesis", by analogy with that of "generation" in compiler construction.

In analysis, the successive phases are :

AM (morphological analysis), obligatory, in ATEF;
AX (expansive analysis X), optional, in EXPANS;
AY (expansive analysis Y), optional, in EXPANS;
AS (structural analysis), obligatory, in ROBRA.

In transfer, the successive phases are :

TL (lexical transfer), obligatory, in EXPANS;
TX (expansive transfer X), optional, in EXPANS;
TS (structural transfer), obligatory, in ROBRA;
TY (expansive transfer Y), optional, in EXPANS.

In generation, the successive phases are :

GX (expansive generation X), optional, in EXPANS;
GS (syntactic generation), obligatory, in ROBRA;
GY (expansive generation Y), optional, in EXPANS;
GM (morphological generation), obligatory, in SYGMOR.

The order of these phases within each step is fixed. Hence, the possible "articulations", all written in TRACOMPL, are AMAX, AMAY, AMAS, AXAY, AXAS, AYAS, ASTL, then TLTX, TLTS, TXTS, TSTY, TSGX, TSGS, TYGX, TYGS, and finally GXGS, GSGY, GSGM and GYGM. As a matter of fact, one needs to write articulations only for composing two phases taken from lingware components using heterogeneous "sets of variables" (see below).

The linguistic operations performed in each phase do not necessarily correspond to their names in a strict manner. For example, morphological analysis may be realized in AM, but it is also possible to distribute it between AM, AX, AY and a fraction of AS (for example, to test for the occurrence of "predicted" possible discontinuous idioms). In general, lexical transfer is also distributed between (at least) TL and TS, for analogous reasons. Similarly, morphological generation of a language such as Arabic [13] may advantageously be distributed between the end of GS and GM.

At the input and output sides of the translation process, the unit of translation is a simple *string of characters.* The 256 EBCDIC characters may be used in the specialized languages to build strings, and all are considered to be atomic (e.g., "é", "É" and "ê" are not known to be variants of "e"). The blank (X'40') is used as separator of *occurrences*. A translation unit, then, is received as a *sequence of occurrences* by the AM phase.

From the output of AM to the input of GM, a unit of translation is represented by a decorated tree. Each phase contains a part, named "DV", where the linguist declares the decoration type, called *set of variables* in Ariane-G5 .

A *decoration,* or *mask of variables* in Ariane-G5 jargon, is a combination of values for all the variables of the considered set, very similar to a property list in LISP.

It is possible to group variables in a hierarchical fashion, the top of the hierarchy being predeclared until a level depending on the specialized language. VAR always denotes the set of variables minus the UL variable. Here is an example :

```
-EXC- ** (key-word for "exclusive").

VSYNTE == (PSYNTE (     ** syntactic exc attr.
   CAT (N, V, A, R, S…),    ** noun, verb… .
   K (PHVB, PHINF, GV, GN, GA))
         ,RSYNTE (  ** syntactic functions.
   FS (SUJ, OBJ1, OBJ2, EPIT, CIRC…)).
VSEME == (PSEME (  ** semantic exc attributes.
   PREDIC (STATE, ACTION, PROC),
   MATTER (DISC, CONT)) ** discrete, concrete.
      ,RS EME (           ** argument roles.
   RL (ARG0, ARG1, ARG2, ARG01, ARG02, TRL10…)
   )).
-NEX- ** non-exclusive attributes.
…
```

A *format* (template) is a constant mask of variables to which a name has been given, in order to use it as an abbreviation in dictionaries and grammars. A *decorated tree* is an oriented and ordered tree where each node bears a decoration.

As in most NLP systems, a lingware written in a specialized language is organized into physically distinct *components,* for reasons of modularity and size, such as variables declarations, formats, procedures, dictionaries and grammars. The components of a phase form an acyclic dependency graph (known by the compiler). A variant of a phase is obtained by selecting some dictionaries & grammars and fixing their priorities.

By combining these choices and the choice of a path in the graph of phases (from AM to GM for a translation), one obtains *execution lines* (for debugging) and *production lines* (for cranking out translations) which are also memorized and managed by Ariane-G5.

### 1.2.3 User interfaces

Ariane-G5 has a complete interactive interface which manages the lingware components as well as the text data base and makes sure that compiled files and intermediate results on texts are consistent with the source lingware at any time [16].

It can also be piloted through the network from the CASH environment developed by E. Blanc in HyperCard on Macintosh [1].

## 2 Inside the French deconverter

### 2.1 Overview

Deconversion is the process of transforming a UNL graph into one (or possibly several) utterance in a natural language. Any means may be used to achieve this task. Many UNL project partners use a specialized tool called DeCo but, like several ather partners, we choose to use our own tools for this purpose.

One reason is that DeCo realizes the deconversion in one step, as in some transfer-based MT systems such as METAL [17]. We prefer to use a more modular architecture and to split deconversion into 2 steps, transfer and generation, each divided into several phases, most of them written in Ariane-G5.

Another reason for not using DeCo is that it is not well suited for the morphological generation of inflected languages (several thousands rules are needed for Italian, tens of thousands for Russian, but only about 20 rules and 350 affixes suffice to build an exhaustive GM for French in Sygmor). Last, but not least, this choice allows us to reuse modules already developed for French generation.

This strategy is illustrated by figure 2.1.

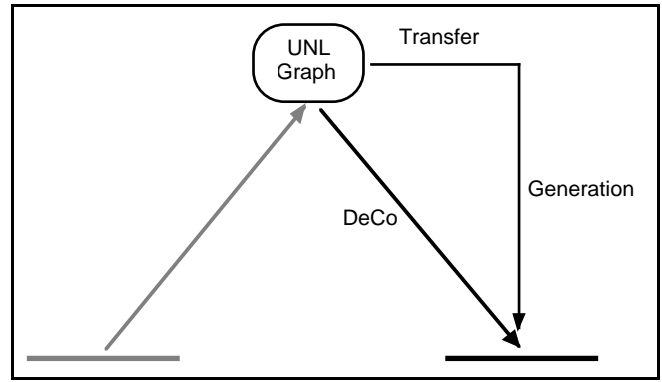

*Fig. 2.1: 2 possible deconversion strategies*

Using this approach, we segment the deconversion process into 7 phases, as illustrated by figure 2.2.

The third phase (graph-to-tree) produces a decorated tree which is fed into an Ariane-G5 TS (structural transfer).

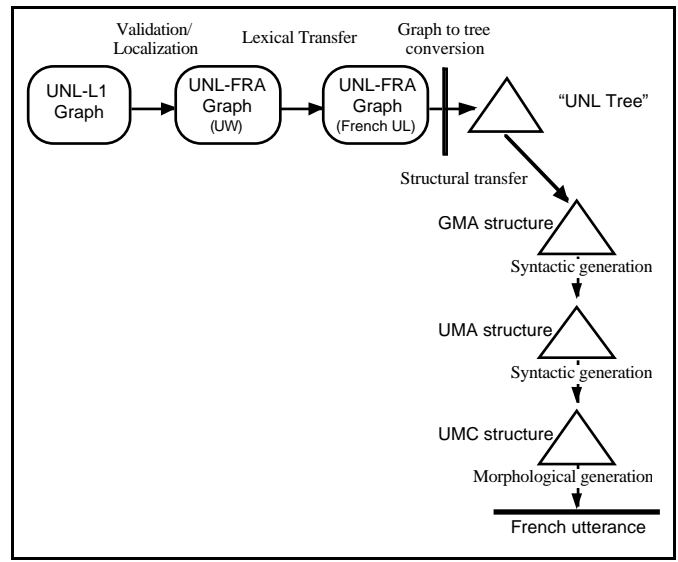

*Fig. 2.2: architecture of the French deconverter*

### 2.2 Transfer

#### 2.2.1 Validation

When we receive a UNL Graph for deconversion, we first check it for correctness. A UNL graph has to be connected, and the different features handled by the nodes have to be defined in UNL.

If the graph proves incorrect, an explicit error message is sent back. This validation has to be performed to improve robustness of the deconverter, as there is no hypothesis on the way a graph is created. When a graph proves valid it is accepted for deconversion.

### 2.2.2 Localization

In order to be correctly deconverted, the graph has to be slightly modified.

#### 2.2.2.1 Lexical localization

Some lexical units used in the graph may not be present in the French deconversion dictionary.

This problem may appear under different circumstances. First, the French dictionary (which is still under development) may be incomplete. Second, the UW may use an unknown notation to represent a known French word sense, and third, the UW may represent a non-French word sense.

We solve these problems with the same method :

Let w be a UW in the graph $G$. Let $D$ be the French dictionary (a set of UWs). We substitute w in $G$ by w' such that : w' $\in D$ and $\forall x \in D$ d(w, w', $G$) $\leq$ d(w, x, $G$). where d is a pseudo-distance function.

If different French UWs are at the same pseudo-distance of w, w' is choosen at random among these UWs (default in non-interactive mode).

#### 2.2.2.2 "Cultural" localization

Some crucial information may be missing, depending on the language of the source utterance (sex, modality, number, determination, politeness, kinship…).

It is in general impossible to solve this problem fully automatically in a perfect manner, as we do not know anything about the document, its context, and its intended usage: FAHQDC[2] is no more possible than FAHQMT on arbitrary texts. We have to rely on necessarily imperfect heuristics.

However, we can specialize the general French deconverter to produce specialized servers for different tasks and different (target) sublanguages. It is possible to assign priorities not only to various parts of the dictionaries (e.g., specialized vs. general), but also to equivalents of the same UW within a given dictionary. We can then define several user profiles. It is also possible to build a memory of deconverted and possibly postedited utterances for each specialized French deconversion server.

### 2.2.3 Lexical Transfer

After the localization phase, we have to perform the lexical transfer. It would seem natural to do it within Ariane-G5, after converting the graph into a tree. But lexical transfer is context-sensititve, and we want to avoid the possibility of transferring differently two tree nodes corresponding to one and the same graph node.

Each graph node is replaced by a French lexical unit (LU), along with some variables. A lexical unit used in the French dictionary denotes a derivational family (e.g. in

---

[2] fully automatic high quality deconversion.

English: destroy denotes destroy, destruction, destructible, destructive…, in French: détruire for détruire, destruction, destructible, indestructible, destructif, destructeur).

There may be several possible lexical units for one UW. This happens when there is a real synonymy or when different terms are used in different domains to denote the same word sense[3]. In that case, we currently choose the lexical unit at random as we do not have any information on the task the deconverter is used for.

The same problem also appears because of the strategy used to build the French dictionary. In order to obtain a good coverage from the beginning, we have underspecified the UWs and linked them to different lexical units. This way, we considered a UW as the denotation of a set of word senses in French.

Hence, we were able to reuse previous dictionaries and we can use the dictionary even if it is still under development and incomplete. In our first version, we also solve this problem by a random selection of a lexical unit.

### 2.2.4 Graph to tree conversion

The subsequent deconversion phases are performed in Ariane-G5. Hence, it is necessary to convert the UNL hypergraph into an Ariane-G5 decorated tree.

The UNL graph is directed. Each arc is labelled by a semantic relation (agt, obj, ben, con…) and each node is decorated by a UW and a set of features, or is a hypernode. One node is distinguished as the "entry" of the graph.

Recall that an ARIANE tree is a general (non binary) tree with decorations on its nodes. Each decoration is a set of variable-value pairs. The graph-to-tree conversion algorithm has to maintain the direction and labelling of the graph along with the decoration of the nodes.

Our algorithm splits the nodes that are the target of more than one arc, and reverses the direction of as few arcs as possible. Here is an example of such a conversion.

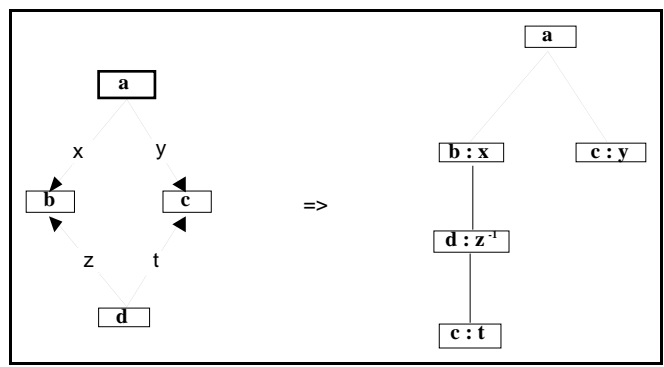

*Fig. 2.3: example graph to tree conversion*

Let $\Sigma$ be the set of nodes of $G$, $\Lambda$ the set of labels, $T$ the created tree, and $N$ is the set of nodes of $T$. The graph $G = \{ (a,b,l) \mid a \in \Sigma, b \in \Sigma, l \in \Lambda \}$ is defined as a set of directed labelled arcs. We use an association list $A = \{ (n_G, n_T) \mid n_G \in \Sigma, n_T \in N \}$, where we memorize the correspondence between nodes of the tree and nodes of the graph.

---

[3] strictly speaking, the same collection of interlingual word senses (acceptions).

```
let  eG ∈ Σ such that e is the entry of G
   eT ← new tree-node(eG,entry)
in T ← eT(); N ← {eT}; A ← {(eG,eT)}
while G ≠ ∅ do
   if there is (a,b,l) in G such that (a,aT) ∈ A then
      G ← G\(a,b,l);
      bT ← new tree-node(b,l);
      A ← A ∪ {(b,bT)};
      let  aT ∈ N such that (a,aT) ∈ A
      in add bT to the daughters of aT;
   else if there is (a,b,l) in G such that (b,bT) ∈ A then
      G ← G\(a,b,l);
      aT ← new tree-node(a,l⁻¹);
      A ← A ∪ {(a,aT)};
      let  bT ∈ N such that (b,bT) ∈ A
      in add aT to the daughters of bT;
   else exit on error ("non connected graph");
```

#### 2.2.5 Structural transfer

The purpose of the structural transfer is to transform the tree obtained so far into a Generating Multilevel Abstract (GMA) structure [4].

In this structure, non-interlingual linguistic levels (syntactic functions, syntagmatic categories…) are underspecified, and (if present), are used only as a set of hints for the generation stage.

### 2.3 Generation

#### 2.3.1 Paraphrase choice

The next phase is in charge of the paraphrase choice. During this phase, decisions are taken regarding the derivation applied to each lexical unit in order to obtain the correct syntagmatic category for each node. During this phase, the order of appearance and the syntactic functions of each parts of the utterance is also decided. The resulting structure is called Unique Multilevel Abstract (UMA) structure.

#### 2.3.2 Syntactic and morphological generation

The UMA structure is still lacking the syntactic sugar used in French to realize the choices made in the previous phase by generating articles, auxiliaries, and non connex compunds such as ne…pas, etc.

The role of this phase is to create a Unique Multilevel Concrete (UMC) structure. By concrete, we mean that the structure is projective, hence the corresponding French text may be obtained by a standard left to right traversal of the leaves and simple morphological and graphemic rules. The result of these phases is a surface French utterance.

## 3 Improving deconversion quality by human interaction

### 3.1 On-line interaction in the current French deconverter

#### 3.1.1 Rationale

Person-system interaction in MT systems has almost exclusively been used for disambiguation, during controlled input (as in [9, 15]), during analysis (on-line mode, as in [10, 12, 18]) or after it (off-line mode, as in [2, 6, 8, 11, 14]). This is because it is felt that, if the intermediate structure produced after analysis or transfer is perfect, a state of the art generator can produce high quality output purely automatically.

In the UNL framework, however, nothing is known about the quality of the input UNL graph, which may have been produced automatically, manually, or interactively. Moreover, some precisions necessary for the target language (sex, aspect, modality, determination…) may not be relevant in the source language and not have been put in the UNL graph.

Finally, although there is a central knowledge base (KB) managed by the UNL Center where all UWs produced by all developers are stored and organized in a gigantic hierarchy along the icl relation and as a network along all other semantic relations, it is unavoidable that there appear at least as many UNL dialects as languages. To improve output quality, it is then necessary to perform some "localization", cultural and lexical.

Because a lot of information is missing, the quality obtainable by an automatic process is inherently limited. Even if the input graph is perfectly correct with regard to the input language, it may be too incomplete and lexically too far away from the target (French) UW set to generate anything but a low quality output. Interaction can be seen as a way to increase the output quality, in a gradual fashion: the more interaction we do, the more precision we get.

In lexical transfer of French UWs into French LUs, the precision of the automatic process is also inherently limited by the strong multilingual character of the architecture.

Suppose for example that we get a UW "chair(icl>furniture)" in a UNL graph coming from an English source. In French and in many other languages, we must choose between a chair with or without arms (fauteuil vs. chaise). The KB may of course contain UWs for these two acceptions, but nothing forces the enconverter to use one of them. Also, in general, the textual context will not contain enough information (such as the phrase "the arms of the chairs…") to trigger a correct lexical selection. But a human has access to the context of previous utterances and to background knowledge, common sense or specialized.

Here again, interaction is the only way to raise output quality.

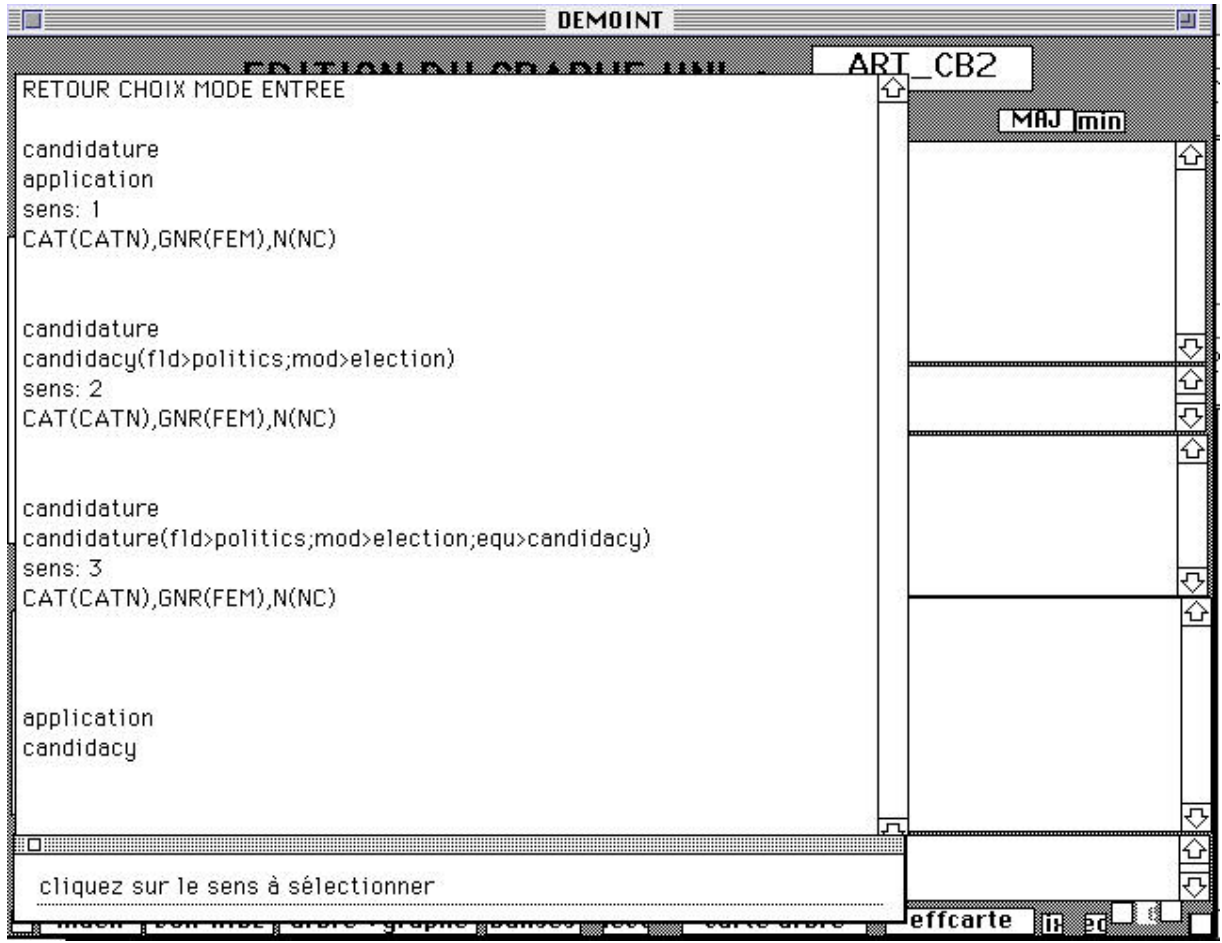

*Fig. 3.1: example of on-line interactive lexical transfer*

### 3.1.2 Current on-line interaction OK for debugging purposes

The current French deconverter can operate in automatic or on-line interactive mode.

In interactive mode, the current UNL graph is shown, together with comments, if any, as comments often contain the English utterance or some explanation in English.

There are two successive possible interactive steps.

- The first handles lexical localization and lexical transfer in a common way: the input UW is shown in a CASH menu, with possible equivalents in the French UW set, and the French lexical units corresponding to each of them. The user selects an appropriate French LU from the menu or forces a better one.

- The second handles cultural localization and consists at this point simply in using a graphical editor to add attributes on some nodes.

As it is organized now, this on-line interaction is clearly reserved for specialists and usable only for development purposes. Figure 3.1 gives an example with the UW "candidature".

Our next goal is to make it usable by naive users, by hiding all technical and specialized aspects.

For instance, we could show only the French LU list, and we could generate dialogue items in standard French to ask questions about sex, modality, aspect, etc. in non-technical terms. But the end user would still feel slave of the system, and perhaps settle for a lower quality to avoid being tied up to the machine while the deconverter works.

That is why we want to investigate the possibility of delaying this interaction after a fully automatic deconversion has been performed, in the same way we have delayed interactive disambiguation after an all-path fully automatic analysis in the LIDIA architecture [2].

### 3.2 Future "active" postedition as off-line interaction with the deconverter

First, it is necessary to link the words of the output text to the nodes in the various trees and graphs they come from, namely, in reverse order, UMC tree, UMA tree, GMA tree, UNL tree, UNL-FRA graph, UNL-L1 graph.

To do this, we will first modify the GS and GM phase to:

- in GS, add one tactical variable containing a canonical index i of each node in the final UMC tree;
- in GM, add to each word or term a special mark such as &i_ corresponding to the leaf node from which it is generated.

Then, in all other phases, we will add another tactical variable containing a canonical index n of each node in the UNL graphs.

A possible strategy is to tackle the easiest parts first. Suppose the posteditor starts the "active learning" preedition mode and selects a word or term. The system uses the implicit links established through the 2 tactical variables and selects the corresponding nodes in each intermediate structure.

#### 3.2.1 Lexical selection during lexical transfer

First, all French LUs corresponding to the selected node will be shown to the posteditor, who will choose one or type one.

This corrected LU will then be put into the corresponding nodes of the UMC, UMA and GMA trees, and the association count of the chosen (UW-FR, LU) pair will be incremented.

On demand, a global replacement operation could then be performed on all the occurrences of the same LU (perhaps in different surface forms) in the utterance.

#### 3.2.2 Lexical localization

If the list shown above does not contain an appropriate LU, the posteditor could ask to enlarge the search.

The system will then go back to the original UNL-L1 tree and search the KB and the French UW dictionary to retrieve the set of possible French-UWs corresponding to the original UW.

It will then propos the union of their possible French LU equivalents, and work as before, with the only difference that the association counts of the chosen (UW-L1, UW-FR) and (UW-FR, LU) pairs is incremented.

#### 3.2.3 Cultural localization

The system will show in another zone an image of the interlingual attributes of the node(s) corresponding to the selected part in the GMA and UMA trees and in the UNL graphs.

A simple menu system will allow the posteditor to change or assign values. The important thing here is to express the corresponding notions in familiar ways.

The same problem will also arise in interactive enconverters. It is encouraging that many previous experiments, in particular in interactive disambiguation for naive users, have shown this to be possible.

#### 3.2.4 New automatic deconversion

After each interaction, the system can produce the corrected French utterance by restarting the deconversion process (in automatic mode) from the modified intermediate structure nearest to the UNL-L1 graph. The interface will simply provide a way for the posteditor to decide whether to redeconvert always, at specific intervals, or on demand.

### 3.3 Open possibilities

We have evoked the possibility of automatically specializing an instance of the deconverter to a given usage (task, sublanguage, user…). Other interesting possibilities should be mentioned here.

#### 3.3.1 Global postediting on a document

First, the UNL document system currently sends deconverters isolated UNL graphs, to produce isolated utterances.

But the document itself is available as an HTML or XML file containing all utterances, and, for each utterance, its UNL graph, its available renderings in some natural languages, and some management information, delimited by UNL tags such as [unl].

The postedition interface could then be built to handle a whole document. It would then become possible to test a correction on one sentence and to apply it then to the rest of the document, as a classical "search-and-replace" operation.

Imagine for instance the benefit of being able to transform all occurrences of "abandon" by "give…up", or all occurrences of "l'information" by "les informations", while taking agreement constraints into account (verb forms will change, etc.).

#### 3.3.2 Style control by GS-oriented interaction

We mentioned earlier the classical assumption that state of the art automatic generators can build high quality outputs from perfect inputs. But that does not imply that these outputs are the *best* for the task at hand.

In the architecture sketched here, it will be quite feasible to introduce interactive style control, by allowing the posteditor to indirectly modify some monolingual attributes on certain nodes in the GMA structure.

The "paraphrase selection" phase (GS1) will take them into account (whenever possible) and regenerate accordingly. We could then, for example, transform all passives in a section by impersonals, etc.

### 3.3.3 Correction of UNL form after negociation with source partner

Last but not least, a very interesting possibility is to modify the original UNL-L1 graph by adding to its nodes all interlingual attributes added to the UNL-FR graph during interactive cultural localization.

The obvious benefit is that this new graph will be more precise and as a result improve automatic deconversion into languages needing the same precisions as French (such as the romance and germanic languages) without any interaction.

Putting it in short, we could say that interactive deconversion might be used to continue the enconversion process to raise the exactness and completeness of the UNL graph wrt some linguistic systems.

But this can not be done arbitrarily, as incorrect modifications by somebody somewhere on Internet might as well degrade the UNL graph. This replacement should be "negotiated" with the enconversion site and/or with some central clearing house, or with the manager of the document system in specialized applications.

## Conclusion

Working on the French deconverter has led to an interesting architecture where deconversion, in principle a "generation from interlingua", is implemented as transfer + generation from an abstract structure (UNL hypergraph) produced from a NL utterance. The idea to use UNL for directly creating documents gets here an indirect and perhaps paradoxical support, although it is clear that considerable progress and innovative interface design will be needed to make it practical.

The other main point of our work is to show how human interaction with a deconverter is both necessary and feasible to raise its output quality. On-line interaction as implemented in the current French deconverter can be used only by specialists, especially for development purposes.

But we have shown how interaction could be delayed and embedded in the postedition phase, which would then interact not directly with the output text, but indirectly with several components of the deconverter. Interacting online or offline can improve the quality not only of the utterance at hand, but also of the utterances processed later, as various preferences may be automatically changed to let the deconverter "learn". Finally, active postedition performed in one target language can be used to enhance the precision of the UNL graph itself, thus indirectly raising the quality of deconversions performed automatically for other languages.

## Acknowledgements

We would like to thank the sponsors of the UNL project, especially UNU/IAS (T. Della Senta) & ASCII (K.Nishi) and of the UNL-FR subproject, especially UJF (C. Feuerstein), IMAG (J. Voiron), CLIPS (Y. Chiaramella), and the French Ministery of Foreign Affairs (Ph. Perez), as well as the members of UNL Center, especially project leader H. Uchida, M. L. Zhu, and K. Sakai. Last but not least, other members of GETA have contributed in many ways to the research reported here, in particular N. Nédeau, E. Blanc, M. Mangeot, J. Sitko, L. Fischer, M. Tomokiyo, and K. Fort.

-o-o-o-o-o-o-o-o-o-o-o-

# Contents

-o-o-o-o-o-o-o-o-o-o-